# A Hierarchical Attention Based Seq2seq Model for Chinese Lyrics Generation

Haoshen Fan[1,2], Jie Wang[2], Bojin Zhuang[2], Shaojun Wang[2] and Jing Xiao[2]

[1] University of Science and Technology of China
[2] Ping An Technology (Shenzhen) Co., Ltd, China
`sa517069@mail.ustc.edu.cn, photonicsjay@163.com,`
`zhuangbojin232@pingan.com.cn, wangshaojun851@pingan.com.cn,`
`xiaojing661@pingan.com.cn`

**Abstract.** In this paper, we comprehensively study on context-aware generation of Chinese song lyrics. Conventional text generative models generate a sequence or sentence word by word, failing to consider the contextual relationship between sentences. Taking account into the characteristics of lyrics, a hierarchical attention based Seq2Seq (Sequence-to-Sequence) model is proposed for Chinese lyrics generation. With encoding of word-level and sentence-level contextual information, this model promotes the topic relevance and consistency of generation. A large Chinese lyrics corpus is also leveraged for model training. Eventually, results of automatic and human evaluations demonstrate that our model is able to compose complete Chinese lyrics with one united topic constraint.

## 1    Introduction

Natural language generation (NLG) (Mann, 1982), also known as text generation, is one of most important tasks in the field of natural language processing (Chowdhury, 2003). NLG has been extensively studied in many applications, such as dialogue system (Chen et al, 2017), machine translation (Cho et al, 2014), text summarization (Nallapati et al, 2016) and so on. In this paper, however, we concentrate on Chinese lyric text generation. Different from prose texts, lyrics exhibits its own significant characteristics, including rhyme, rhetoric and repeated structures. In the perspective of narrative, a lyrics paragraph always concentrates on one main topic due to its limited length, which is totally different from long documents often covering several topics. Moreover, sentence length of lyrics is always short, in a range of 8 to 15 words, which results in close contextual relationship between adjacent sentences.

In general, most of text generation models can be extended for lyrics generation. In the area of text generation, there exist two main approaches, one of which is probabilistic language model (LM) and the other is Sequence-to-Sequence (Seq2Seq). LM has been successfully used in various NLG applications, which is capable of predicting next words on the premise of prior contexts. For instance, Bengio used n-gram model of three layers to construct a language model (Bengio et al, 2003). Then, Mikolov promoted LM with recurrent neural network (RNN) (Mikolov, 2010). However, LM even with long-short term memory (LSTM) network would suffer from semantic shift along with the accumulation of sequence length (Hochreither and Schmidhuber, 1997). To address sequence transduction between heterogeneous data, a sequence-to-sequence model was proposed (Sutskever et al, 2014). Taking a sequence as input, Seq2Seq can encode it into a fixed dense vector and then decode to another sequence. Moreover, Bahdanau applied the attention mechanism to the Seq2Seq in order to diffuse decoding weights into different parts of input (Bahdanau et al, 2015). Based on Seq2Seq, text generation can be defined as next sentence prediction on the premise of prior sentences. In most of Seq2Seq applications, however, input contexts are formed based on sequential concatenation of previous sentences directly. Consequently, the semantic effect of sub-sequences far from the decoder could become weaker on prediction.

To generate long-paragraph Chinese lyrics with high contextual relevance and consistence, in this paper, we propose a hierarchical recurrent encoder (HRE) incorporated into the seq2seq framework. HRE can extract both sentence-level and word-level semantics from prior sentences, providing more contextual information for decoding. Moreover, the attention mechanism covering the most adjacent sentence is applied, considering the closest connection with next prediction. The rest of the article will be structured as follows: Section 2 describes the data preprocess of Chinese lyrics corpus, Section 3 describes the details of our model, Section 4 describes the experiments on several models, Section 5 briefly introduces the related work and we make some conclusion in Section 6.

## 2   Related Work

NLG is an essential part of natural language processing (NLP). According to the modality of input, there exist text-to-text generation, meaning-to-text generation, data-to-text generation, image-to-text generation etc. In this paper, lyrics generation is modeled as a specific text-to-text generation with previous sentences as input. Similar tasks including Chinese poetry generation (Wang et al, 2016), essay generation (Feng et al, 2018) and comment generation (Tang et al, 2016) have been extensively studied. Chinese poetry generation generate a kind of hierarchical text with strict format which often has a fixed number of sentences and each sentence has a fixed number of words. For instance, to generate context-aware comments, Tang proposed to encode the context as a continuous semantic representation into a basic RNN model. Moreover, essay generation covering several topic words has also been demonstrated by similar methods.

Various hierarchical models have been used for generating coherent long texts. For example, Li proposed a hierarchical neural auto-encoder to build an embedding for a paragraph (Li et al, 2015). Lin presented a novel hierarchical recurrent neural network language model (HRNNLM) to maintain overall coherence in a document (Lin et al,

2015). Following the HRED proposed by Sordoni (Sordoni et al, 2015), Serban extended the hierarchical model to promote dialogue generation with long-term contexts (Serban and Bengio et al, 2016). Later, he enhanced the HRED model with a latent variable at the decoder (Serban and Sordoni et al, 2016). Furthermore, a hierarchical seq2seq with attention is proposed by us for Chinese lyrics generation to address the long-term coherence.

## 3    Model

In this section, a hierarchical attention based Seq2Seq model for lyrics generation is described. Original lyrics has been preprocessed into the paragraph format for model training in advance. Here, a lyrics paragraph comprises a sequence of $M$ sentences, i.e. $P = \{S_1, S_2, ..., S_M\}$. Each sentence $S_m$ consists of a sequence of $N_m$ words $S_m = \{\omega_{m,1}, \omega_{m,2}, ..., \omega_{m,N_m}\}$, where $\omega_{m,n}$ represents the word at position n in sentence m.

### 3.1    Recurrent Neural Network

A recurrent neural network (RNN) model recurrently calculates a vector named recurrent state or hidden state $h_n$ by taking a sequence of words $\{\omega_1, \omega_2, ..., \omega_N\}$:

$$h_n = f(h_{n-1}, \omega_n), n \in (1, N), h_0 = 0 \qquad (1)$$

Particularly, the $h_0$ denote the initial state and always is set as zero at the time of training. Usually, $h_n$ depends on the current word $\omega_n$ and previous ones before the current time step. In **Equation 1**, $f$ denotes a parametrized non-linear function, such as sigmoid, hyperbolic tangent, long-short term memory (LSTM) and gate recurrent unit (GRU). The hidden state will lose long contextual information when a vanilla RNN such as sigmoid or hyperbolic tangent is used. Through bringing in a memory cell, LSTM or GRU can handle longer-term contexts. Moreover, GRU requires less computational cost compared with LSTM. Thus, GRU is used as the RNN cell unit. The equations of GRU are summarized as follows:

$$z_t = \sigma(W_z \omega_t + U_z h_{t-1}) \qquad (2)$$

$$r_t = \sigma(W_r \omega_t + U_r h_{t-1}) \qquad (3)$$

$$\tilde{h}_t = \tanh(W\omega_t + U(r_t * h_{t-1})) \qquad (4)$$

$$h_t = (1 - z_t) * h_{t-1} + z_t * \tilde{h}_t \qquad (5)$$

In the Equation above, the $\sigma$ is the non-linear fuction i.e. logistic sigmoid, which limits output to range [0,1]. $z_t$ is the update gate deciding the weight of input information past, and $r_t$ is the reset gate determining the weight of last state. The candidate update $\tilde{h}_t$ controls the percentage of information obtained from $h_{t-1}$ with reset gate. The final update $h_t$ depends on the update gate and candidate update. The subscript letter $t$ represents the time step.

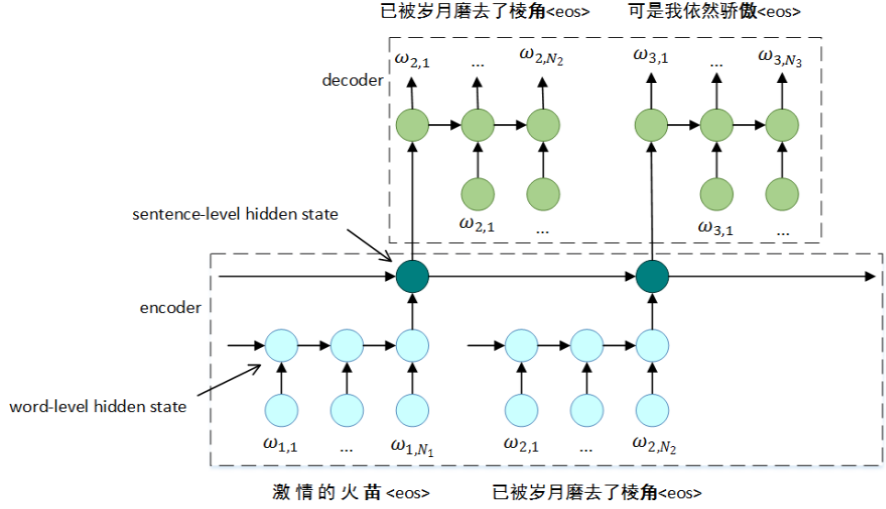

**Fig. 1.** The graph of HRED model constructing a Chinese lyrics paragraph with three sentences. The word-level encoder the sentences into a fix dense vector and the sentence-level encoder map the vectors into the representation of paragraph, which is the input of decoder. We bold the last rhyming word.

### 3.2 Hierarchical Recurrent Encoder

Sordoni proposed a hierarchical recurrent encoder-decoder (HRED) to predict a next web query conditioned on previous queries submitted by users (Sordoni et al, 2015). The hierarchical encoder consists of query-level and session-level encoders, which has been demonstrated very successful for web query prediction. Following this HRED work, a lyrics paragraph is considered with hierarchical structure of word-level and sentence-level as shown in **Fig. 1**. At the bottom of the network, the sentence-level RNN encodes each sentence into a fix dense vector. This higher-level semantic vector is used to predict the next sentence $S_{m+1}$.

Different from web queries, however, a lyrics paragraph always contains more than ten sentences. Thus, we adapt this HRE to handle a certain number of sentences before decoding as shown in **Fig. 2**. The number of sub-group sentences is denoted as $Num$, which is a hyper-parameter. After some trial and error, the $Num$ is optimized as 5. Note that GRU is used as the basic RNN cell unit. Moreover, the word-level encoder and the decoder share same parameters.

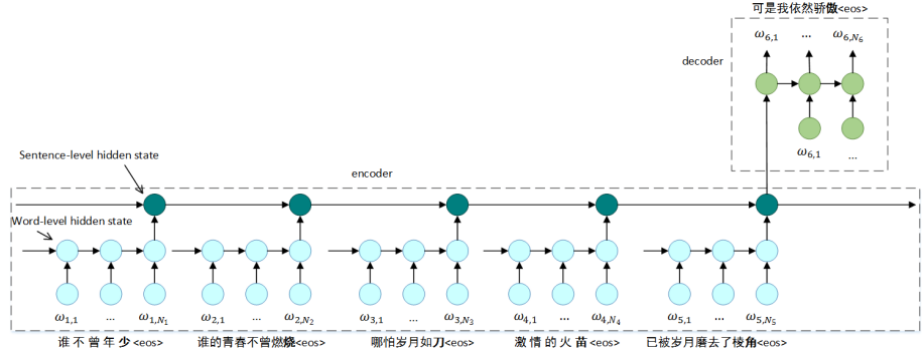

**Fig. 2.** Hierarchical Seq2Seq extending HRED.

### 3.3 Decoder

In the decoder, the last state of the sentence-level RNN is used as the initial state. The probability distribution in the time t is represented:

$$p(\omega_t|s, \omega_1, \ldots, \omega_{t-1}) = g(h_{t,dec}, \omega_{t-1}, s) \quad (6)$$

In the **Equation 6**, the $s$ is the last state of the sentence-level encoder. The state $h_{t,dec}$ can be denoted as:

$$h_{t,dec} = f(h_{t-1,dec}, \omega_{t-1}, s) \quad (7)$$

Seq2Seq with attention was first proposed by Dzmitry and has achieved a great success in various NLG applications. Here, the attention mechanism is incorporated into the hierarchical model and applied to the word-level encoder. The difference between seq2seq with attention and conventional seq2seq is that the decoder uses different context vector $s_t$ in each step as:

$$h_{t,dec} = f(h_{t-1,dec}, \omega_{t-1}, s_t) \quad (8)$$

The context vector $s_t$ is a weighted sum of the encoder hidden states { $h_{1,dec}$, $h_{2,dec}, \ldots, h_{N_m,dec}$ }:

$$s_t = \sum_{j=1}^{N_m} a_{tj} h_{j,enc} \quad (9)$$

where the $a_{tj}$ is computed by decoder hidden state $h_{t-1,dec}$ and each encoder hidden state { $h_{1,dec}, h_{2,dec}, \ldots, h_{N_m,dec}$ }. As shown in **Fig. 3**, we only use the sequence of hidden states of last sentence $S_{m-1}$ as the input of attention while predicting the next sentence $S_m$ because of the strongest semantic relationship between adjacent sentences. Finally, beam search is used in the inference stage.

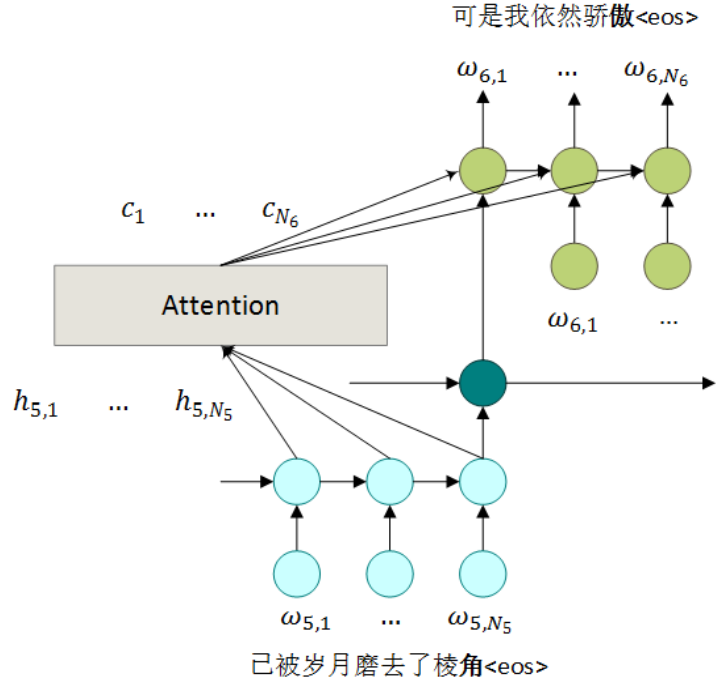

**Fig. 3.** Hierarchical Seq2Seq with attention.

## 4      EXPERIMENTS AND RESULTS

In this section, settings of experimental parameters are described at detail. A generic Seq2Seq model is applied as a baseline. Tensorflow framework is used to implement the hierarchical attention based Seq2Seq model because of its flexibility and accumulated development experiences shared in community (Tang, 2016).

### 4.1     DATA PROCESSING

Lyrics in monolingual Chinese was collected to guarantee the same data structure. 100,000 pop song lyrics has been prepared, which is familiar with most Chinese Netizens. Based on this corpus, a prior vocabulary of 7030 words was achieved. Filtering out 1985 low frequency words which occur less than 10 times in the paragraphs, our vocabulary size is eventually 5045. Additionally, the following three symbols have been added into this vocabulary, including 'unk' representing unknown words, 'go' and 'eos' donating the start and end of sentences. Besides, the maximum length of all sentences is limited to 20. Those sentences longer than 20 have been filtered out. Finally, the prepared corpus was divided into two parts, 90% as training data while 10% as test data.

### 4.2 Parameters Setting

We use the word embedding with dimension 300 to represent the words. Specifically, the word embedding is defined as the trainable parameters, which will be fine-tuned as the training progress. The word-level encoder has 1000 hidden unit. To keep the sentences talking the same topic and memorizing complex topics and emotion, we set the dimensionality of sentence-level encoder and decoder to 1500. Moreover, the word-level encoder has 3 layers to ensure the model can encode the complex lyrics sentences while the sentence-level encoder and decoder has 1 layer. Finally, the beam width k is set to 5. All of the parameters are randomly initialized within the range [-0.5,0.5]. They are trained to minimize the cross-entropy loss function with the Adam optimizer (Kingma and Ba, 2015). We set the mini-batch to 256. We train the model until the loss function has a minimum value and is no less than that in the next three epochs.

### 4.3 Evaluation Metrics

**Human Evaluation**
Nine Chinese experts are asked to evaluate the performance of our model. They are asked to mark generated lyrics samples from three different aspects: Topic Relevance, Fluency and Semantic Coherence. The score is range from 1 to 5. 5000 lyrics paragraphs are randomly generated for graduate students majored in Music to score.

**BLEU**
Additionally, we use Bilingual Evaluation Understudy (BLEU) as our automatic evaluation (Papineni et al, 2002). BLEU is an evaluation method widely used for machine translation. In this paper, the test dataset is used as the reference ground truth for automatic evaluation.

**Table 1.** Averaged score of different model for lyrics text generation.

| model | Topic Relevance | Fluency | Semantic Coherence | Average Score |
|---|---|---|---|---|
| Seq2Seq | 2.34 | 2.99 | 2.38 | 2.57 |
| Hierarchical Seq2Seq | 2.99 | 2.79 | 2.76 | 2.85 |
| **Hierarchical Seq2Seq with attention** | **3.11** | **3.17** | **3.46** | **3.24** |

### 4.4 Experimental Results

**Table 1** shows the final results of human evaluation of different models. The basic Seq2Seq model exhibits the worst performance since it only considers the adjacent sentences, which can't maintain the long-term semantic coherence. In comparison, the hierarchical Seq2Seq model boosts the performance in terms of "Topic Relevance" and "Semantic Coherence". The main reason is that the hierarchical model is able to remember higher-level semantics due to the sentence-level encoding. However, the

8poor performance of the hierarchical model in "Fluency" is attributed to the omission of word-level encoding. Thus, the hierarchical Seq2Seq with attention performs best in all three perspectives. The attention mechanism helps the model directly connect the semantic relationship between adjacent sentences while retaining higher-level contextual information.

In order to make the evaluation result more objective, we also show the BLEU result in **Table 2**. Obviously, the results of BLEU show the same trend as those of human evaluation. The hierarchical Seq2Seq preforms better than Seq2Seq model and the hierarchical Seq2Seq with attention performs better than the hierarchical Seq2Seq. Compared with other area of text generation such as machine translation, the BLEU results are very small. The reason is that the generated lyrics use different word combinations to express the same meaning while each word of the text to be translated often has a unique correct answer. Finally, a sample of generated lyrics is given in **Table 3**. Those underlined and bold Chinese characters at the ending of sentences are rhyming.

**Table 2.** BLEU scores of different models.

| model | BLEU Score |
|---|---|
| Seq2Seq | 0.189 |
| Hierarchical Seq2Seq | 0.274 |
| Hierarchical Seq2Seq with attention | 0.288 |

**Table 3:** Example of generated lyrics. The blue text is the lyrics generated by the hierarchical Seq2Seq model with attention. Note that the first line "Homeland" is the title of the lyrics.

*故乡*
*Homeland*

看那田地看那原野，
Look at the farmland and look at the field,
一片美丽好风**光**，
It's a beautiful scenery.
俄罗斯的大自然啊，
Nature in Russia,
这是我的故**乡**。
Is my hometown.
看那高山看那平地，
Look at the mountain and the at land,
无边草原和牧**场**。
Borderless grasslands and pastures.
俄罗斯的辽阔地**方**，
Russia's vast territory,
是我梦中的故**乡**。

99

<div style="text-align:center">

*Is my dream home.*
看那远方的山，
*Look at the mountains in the distance,*
看那辽阔的草原，
*Look at the vast grassland,*
是那遥远的天**堂**，
*It's a distant paradise,*
那遥远的故**乡**。
*The remote homeland.*

</div>

## 5 Conclusions

In this paper, we propose a novel hierarchical Seq2Seq model with attention for Chinese lyrics generation. A large-scale Chinese lyrics corpus has been leveraged for model training. Results of human and BLEU evaluation demonstrate the effectiveness of this model owing to its sentence-level semantic encoding and attended to adjacent sentences. Moreover, this hierarchical encoder method offers a promising approach of context fusing for other NLG applications.

## 6 Acknowledgement

This work was supported by Ping An Technology (Shenzhen) Co., Ltd, China.